\documentclass{article}

\PassOptionsToPackage{numbers,compress}{natbib}

\usepackage[preprint]{neurips_2026}

\usepackage[utf8]{inputenc}
\usepackage[T1]{fontenc}
\usepackage{hyperref}
\usepackage{url}
\usepackage{booktabs}
\usepackage{graphicx}
\usepackage{amsfonts}
\usepackage{amsmath}
\usepackage{amssymb}
\usepackage{nicefrac}
\usepackage{microtype}
\usepackage{xcolor}

\title{Sharpen Before You Adapt: Data-Free Entry-State\\
Sharpening for Test-Time Reinforcement Learning}

\author{%
  Zhanming Zhang\\
  Independent Researcher\\
  New York, NY\\
  \texttt{zmzhang15@gmail.com}\\
  \And
  Vinoth Selvendran\\
  Independent Researcher\\
  Palo Alto, CA\\
  \texttt{vinoth459005@gmail.com}\\
}

\begin{document}

\maketitle

\begin{abstract}
Test-time reinforcement learning (TTRL) adapts language models on unlabeled test problems using supervision derived from their own samples. This makes the checkpoint's \emph{entry state} consequential: a diffuse policy provides noisier self-supervision and may spend much of a limited adaptation budget merely concentrating probability mass before reliably expressing capability it already possesses. We propose \textbf{entry-state sharpening}: use data-free training \emph{before} TTRL to prepare a general-purpose checkpoint in a state that subsequent label-free adaptation can exploit more efficiently. The idea is not tied to one training recipe; different data-free objectives can move the same base model to different entry states. Across five data-free checkpoints derived from Qwen3-4B and evaluated under an identical 15-step TTRL protocol, entry policy entropy strongly rank-orders endpoint conversion efficiency, a reliability-to-reachability measure (Spearman $\rho=-0.90$; $\rho=-0.99$ after controlling for entry reachability). The contrast across objectives is striking: R-Zero remains diffuse at $3.39$ nats and finishes below the untuned base in 6/6 matched comparisons across MATH, GPQA, and AMC, whereas SPIRAL reaches $0.07$ nats and achieves the highest post-TTRL accuracy on MATH and GPQA despite its self-play stage using no math training data. An in-domain label-free self-distillation intervention further shows that the entry state can be deliberately sharpened. These results motivate treating checkpoint preparation as a \emph{state-control problem}: use data-free training to improve TTRL readiness, with entry entropy as a label-free control signal and reachable capability as the constraint.

\end{abstract}


\section{Introduction}
\label{sec:intro}

Test-time reinforcement learning (TTRL)~\cite{ttrl} adapts a language model directly on unlabeled test problems, using agreement among the model's own samples as supervision~\cite{selfconsistency}. This makes TTRL attractive when labels or verifiers are unavailable: a model can continue improving from a few hundred target problems and a small number of optimizer steps.

But TTRL has an implicit prerequisite that is rarely treated as a design variable: the \emph{state of the checkpoint when adaptation begins}. Its learning signal is generated from the checkpoint's own distribution. If that distribution is diffuse, two problems arise simultaneously: the model's self-consensus is less reliable, and part of the limited adaptation budget must first be spent concentrating probability mass before reachable capability can be expressed consistently. A checkpoint can therefore be strong at frozen inference yet poorly positioned for further self-improvement.

We propose \textbf{entry-state sharpening}: use data-free training \emph{before} TTRL to prepare a general-purpose checkpoint in a state that subsequent label-free adaptation can exploit more efficiently. The proposal is deliberately broader than a particular training recipe. Self-play, self-evolution, and self-distillation can all reshape the policy handed to TTRL; the key question is what \emph{entry state} they produce. We therefore treat checkpoint preparation as a state-control problem rather than as a choice of upstream recipe.

This view follows from evidence that reinforcement learning with verifiable or self-generated rewards often improves small-sample performance by re-weighting behaviors already accessible to the base model~\cite{yue2025rl,invisibleleash,sharpening}. In this regime, TTRL is largely a \emph{conversion} process: turning reachable capability into reliable output. To separate these quantities we define conversion efficiency $\eta=\mathrm{mean@16}/\mathrm{best@16}$, where mean@16 measures sampled accuracy and best@16 estimates reachability (formalized in \S\ref{sec:method}). A useful entry state should raise this conversion efficiency without reducing what the model can still reach.

We study five data-free checkpoints derived from the same Qwen3-4B base~\cite{qwen3}, each evaluated under an identical 15-step TTRL protocol. Across these checkpoints, \textbf{entry policy entropy} strongly rank-orders endpoint conversion efficiency: Spearman $\rho=-0.90$, strengthening to $\rho=-0.99$ after controlling for entry reachability. The ordering is also stable across seeds ($\rho=1.00$, $p=0.017$), whereas raw-accuracy ordering is not. Entry entropy therefore provides a label-free readout of a checkpoint property closely associated with its ability to convert capability under TTRL.

The contrast between training objectives makes the entry-state problem concrete. R-Zero~\cite{rzero} enters highly diffuse at $3.39$ nats and finishes below the untuned base in all 6/6 matched comparisons across MATH, GPQA, and AMC. SPIRAL~\cite{spiral}, despite never training on math, enters at $0.07$ nats and yields the strongest post-TTRL checkpoint. Thus data-free training is not inherently preparatory: different objectives can move the same base family toward very different states for subsequent adaptation.

We further show that this state can be manipulated directly. Our in-domain label-free self-distillation operator moves the Base checkpoint from $1.186$ to $0.183$ nats and increases endpoint conversion efficiency from $0.941$ to $0.945$. At this high preparation strength, however, final reachability falls from $0.820$ to $0.811$. Entry-state sharpening is therefore not entropy minimization; its goal is to improve conversion while preserving reachable capability.

Our contributions are:
\begin{itemize}
\item \textbf{Entry-state sharpening.} We formulate checkpoint preparation for TTRL as a data-free state-control problem: reshape the model before adaptation so that its existing capability can be converted more efficiently.
\item \textbf{A measurable entry-state signal.} Across five data-free checkpoints, entry policy entropy strongly rank-orders endpoint conversion efficiency, including after controlling for entry reachability.
\item \textbf{Evidence that preparation objectives matter through the states they produce.} R-Zero and SPIRAL move the same base family to opposite ends of the entropy axis and exhibit correspondingly different TTRL behavior.
\item \textbf{A direct in-domain intervention.} Label-free self-distillation substantially sharpens the entry state and improves conversion, while its high-strength endpoint identifies reachability as the constraint on further sharpening.
\end{itemize}

\section{Related Work}
\label{sec:related}

\textbf{Test-time adaptation and label-free self-improvement.}
Test-time training adapts models on unlabeled test inputs using self-supervised objectives~\cite{ttt}. TTRL~\cite{ttrl} brings this idea to language-model reasoning by optimizing a majority-vote reward with GRPO~\cite{deepseekmath}; related methods instead use model confidence, entropy, or self-certainty as the adaptation signal~\cite{rent,emeffective,oneshotem,empo,intuitor,selfcertainty}. Recent work also regulates uncertainty \emph{within} the adaptation loop to avoid unreliable pseudo-labels or premature collapse~\cite{spine,ettrl,echo,scrl,ttrlcocov}. We study a complementary question: rather than modifying TTRL itself, what state should the checkpoint be in \emph{before} adaptation begins? This distinction is motivated by evidence that RL primarily re-weights behaviors already accessible to the base policy~\cite{yue2025rl,invisibleleash,sharpening}.

\textbf{Data-free self-evolution and preparation for downstream RL.}
A growing family of methods improves language models without external labeled data, including uncertainty-driven co-evolution in R-Zero~\cite{rzero} and its multi-solver disagreement extension~\cite{multisolver}, verifier-grounded self-play in Absolute Zero~\cite{azr}, self-instruction in SeRL~\cite{serl}, and game self-play in SPIRAL~\cite{spiral}. Label-free self-training methods similarly fine-tune models on their own selected generations~\cite{lmsi,rft,rest}. These works primarily evaluate the resulting checkpoint as an endpoint; we instead study the checkpoint as an \emph{initialization for further test-time learning}. This connects to a broader line on preparing models for reinforcement learning: domain-adaptive pretraining and mid-training can substantially affect downstream RL performance and scalability~\cite{dapt,olmo2,octothinker,rmt,gandhi,spurious}. SPIRAL is particularly close: inserting self-play before RLVR improves over RLVR alone~\cite{spiral}. Our setting differs in that the downstream learner is itself label-free and operates on the test distribution, and we characterize preparation through a measurable property of the resulting checkpoint rather than through the upstream recipe alone.

\textbf{Concentration, entropy, and coverage.}
Self-training and distillation are known to concentrate output distributions and, when pushed far enough, reduce multi-sample diversity or coverage~\cite{gem,osdiversity,losecoverage}. This motivates treating sharpening as a controlled preparation step rather than unconditional entropy minimization. At the same time, long-horizon RL literature often treats entropy collapse as a failure mode and explicitly preserves exploration~\cite{entropy_mechanism,dapo,forking,prorl}. Our regime is different: TTRL uses a short adaptation horizon and self-generated supervision, where entry concentration can improve the reliability and efficiency of subsequent updates. We therefore distinguish \emph{entry-state concentration} from global entropy collapse, and study how much concentration is useful before reachable capability begins to erode.

\section{Method: Entry-State Sharpening}
\label{sec:method}

Let $\pi_0$ be a general-purpose checkpoint and $\mathcal{A}_T$ a fixed $T$-step test-time adaptation procedure on unlabeled target problems $\mathcal{X}$. Standard TTRL adapts directly from $\pi_0$:

\[
    \pi_T=\mathcal{A}_T(\pi_0;\mathcal{X}).
\]

We instead insert a data-free preparation stage,

\[
    \pi_{\mathrm{entry}}
    =
    \mathcal{P}_m(\pi_0;\tau_m),
    \qquad
    \pi_T
    =
    \mathcal{A}_T(\pi_{\mathrm{entry}};\mathcal{X}),
\]

where $m$ denotes a preparation objective and $\tau_m$ its method-specific training strength. We call this framework \textbf{entry-state sharpening}.

The preparation operator is not tied to one algorithm. Self-play, self-evolution, and self-distillation can all reshape the checkpoint before TTRL; what matters is the \emph{state} they produce. Their training strengths are not directly comparable---e.g., self-play iterations and self-distillation steps measure different quantities---so we compare them through a common checkpoint-level readout.

We separate reachable capability from how reliably it is expressed using

\[
    \eta
    =
    \frac{\mathrm{mean@16}}{\mathrm{best@16}},
    \qquad
    \mathrm{mean@16}
    =
    \eta\cdot\mathrm{best@16},
\]

where mean@16 is sampled accuracy and best@16 estimates reachability. Entry-state sharpening therefore aims to improve subsequent conversion efficiency $\eta$ while preserving reachability.

Our common state variable is \textbf{entry policy entropy}. Let $x$ be a problem from the first TTRL batch, $z=(z_1,\ldots,z_L)$ a sampled reasoning trace, $z_{<t}$ its prefix at token $t$, and $\mathcal{V}$ the vocabulary. We measure

\[
H_{\mathrm{entry}}(\pi)
=
\mathbb{E}_{x,z}
\left[
\frac{1}{|z|}
\sum_{t=1}^{|z|}
-\sum_{v\in\mathcal V}
\pi(v\mid x,z_{<t})
\log\pi(v\mid x,z_{<t})
\right]
\]

before the first TTRL update. It requires no ground-truth labels and is comparable across preparation methods. We compute $H_{\mathrm{entry}}$ on the checkpoint's own rollouts from the first TTRL training batch, using TTRL's prompt template, tokenization, and rollout temperature.

We study three data-free routes to different entry states. We first consider two representative paradigms of data-free self-evolution. Game-based self-play generates experience through interaction and learns from environment-verified outcomes; we use \textbf{SPIRAL}~\cite{spiral} as a representative instance. Curriculum-based self-evolution instead generates or selects problems according to the model's current behavior and trains on self-generated supervision; we use \textbf{R-Zero}~\cite{rzero}, whose uncertainty-driven curriculum targets problems on which the Solver disagrees with itself. These two paradigms provide a useful contrast because both require no labeled target-domain data, yet their objectives exert very different pressures on the resulting policy. Finally, we construct an \textbf{in-domain self-distillation} checkpoint as a direct sharpening intervention. Before TTRL, the base model generates reasoning traces on unlabeled target problems; traces consistent with its self-consistency majority are retained as pseudo-demonstrations and used for supervised cross-entropy fine-tuning~\cite{lmsi,rft,rest}. We use self-distillation here as one concrete operator for deliberately moving the entry state.

\section{Experimental Setup}
\label{sec:setup}
\label{sec:metrics}

All checkpoints derive from Qwen3-4B~\cite{qwen3}. We evaluate them under one frozen TTRL protocol: 15 optimizer steps; validation at steps $0,5,10,15$; 32 reward votes; 16 rollout/update samples and 16 validation samples per question; and matched seeds. No ground-truth labels enter adaptation. The questions used for training are identical across the checkpoints.

Our primary roster contains five checkpoints, all produced without labeled target answers: Base; R-Zero~\cite{rzero}; SPIRAL@400~\cite{spiral}, the public multi-game SPIRAL self-play release (a completed run at SPIRAL's nominal 400-step budget); SPIRAL@32, an early step-32 endpoint of a Kuhn-Poker self-play run; and our in-domain self-distilled checkpoint.

For each validation question, we report mean@16, best@16, and conversion efficiency $\eta$. best@16 is a 1,000-resample bootstrap estimate of reachability rather than the literal fraction of questions with one observed correct response. Entry best@16 is used as the capability control in partial-correlation analyses.

\textbf{Training configuration.}
TTRL optimizes the majority-vote reward with GRPO~\cite{deepseekmath} using \texttt{verl}. Each of the 15 updates uses 32 unlabeled problems. The actor learning rate is $5\times10^{-7}$ with cosine decay. Rollouts are generated with vLLM at temperature $1.0$, while validation uses 16 samples per problem at temperature $0.6$ and top-$p=0.95$. Each run uses eight NVIDIA A100 40\,GB GPUs.

\section{Results}
\label{sec:results}

\subsection{Frozen accuracy does not imply TTRL readiness}
\label{sec:crossover}

The motivation for entry-state sharpening is that a checkpoint's static capability and its ability to continue learning can diverge.

\begin{figure}[t]
\centering
\includegraphics[width=\linewidth]{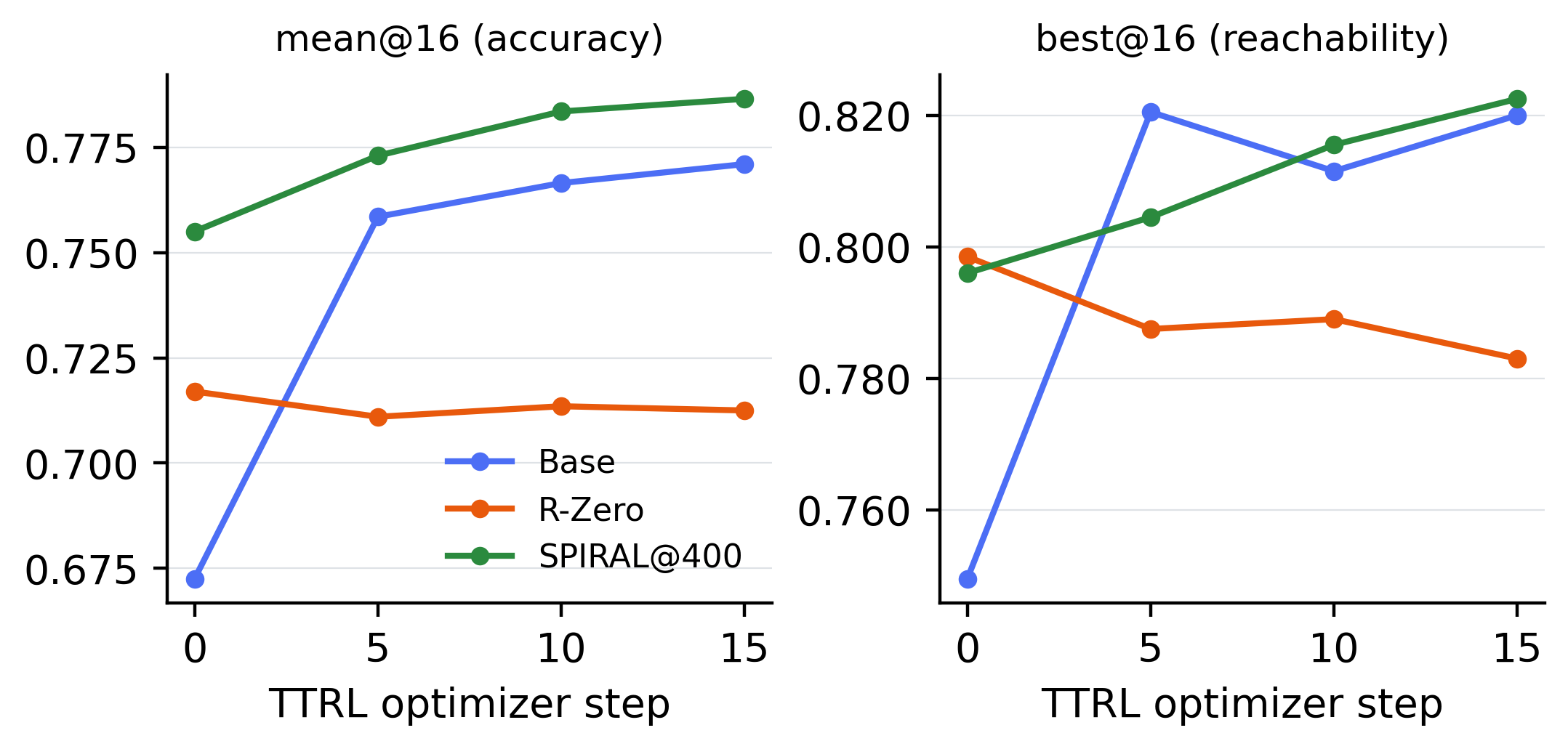}
\caption{Identical 15-step TTRL on unlabeled MATH problems for Base, R-Zero, and SPIRAL@400; means over seeds $\{2,3\}$. R-Zero starts $4.5$ accuracy points ahead of Base and finishes $5.8$ points behind it, while SPIRAL@400 finishes highest despite never training on math.}
\label{fig:traj}
\end{figure}

The three focal checkpoints enter MATH with similar reachability but sharply different entry entropy. Entry best@16 ranges from $0.749$ to $0.799$, whereas entropy ranges from $0.073$ nats for SPIRAL@400 to $3.391$ for R-Zero; Base lies between them at $1.186$.

Figure~\ref{fig:traj} shows that frozen-score ordering does not survive adaptation. R-Zero begins above Base in both accuracy ($0.717$ vs.\ $0.672$) and reachability, but after 15 steps finishes below it ($0.713$ vs.\ $0.771$), with lower conversion efficiency ($0.911$ vs.\ $0.941$). SPIRAL@400 shows the opposite pattern: despite never training on math, it reaches the highest final accuracy ($0.787$) and conversion efficiency ($0.957$).

The inversion is not confined to MATH. Under matched seeds, R-Zero finishes below Base on final mean@16 in all three domains (Tables~\ref{tab:zoo}--\ref{tab:amc}). These results show that frozen capability alone does not identify which checkpoint is a good substrate for TTRL.

\subsection{Entry entropy rank-orders test-time conversion}
\label{sec:zoo}

Entry-state sharpening makes a more specific prediction than ``lower entropy gives higher accuracy.'' Final accuracy depends on both reachable capability and how reliably it is expressed. We therefore ask whether entry concentration predicts conversion efficiency.

\begin{table}[t]
\centering
\caption{The five data-free checkpoints under the frozen 15-step TTRL protocol on MATH, ordered by entry entropy. Entry columns are seed means; final columns and $\eta$ are means $\pm$ sample s.d.\ over seeds $\{2,3\}$. Endpoint conversion efficiency $\eta$ is strongly ordered by entry entropy.}
\label{tab:zoo}
\scriptsize
\setlength{\tabcolsep}{3pt}
\begin{tabular}{llcccccc}
\toprule
\textbf{Checkpoint} & \textbf{Objective} & \textbf{Entry} & \textbf{Entry} & \textbf{Entry} &
\textbf{Final} & \textbf{Final} & \boldmath$\eta$ \\
& & \textbf{entropy} & \textbf{mean@16} & \textbf{best@16} &
\textbf{mean@16} & \textbf{best@16} & \\
\midrule
R-Zero         & uncert.\ curriculum    & $3.391$ & $0.717$ & $0.799$ & $0.713 \pm 0.004$ & $0.783 \pm 0.007$ & $0.911 \pm 0.013$ \\
Base           & --- (pretrained)       & $1.186$ & $0.672$ & $0.749$ & $0.771 \pm 0.005$ & $0.820 \pm 0.006$ & $0.941 \pm 0.001$ \\
SPIRAL@32      & game play, Kuhn @32    & $0.344$ & $0.747$ & $0.806$ & $0.771 \pm 0.015$ & $0.823 \pm 0.013$ & $0.937 \pm 0.003$ \\
Self-distilled & data-free, high-dose   & $0.183$ & $0.710$ & $0.771$ & $0.767 \pm 0.016$ & $0.811 \pm 0.018$ & $0.945 \pm 0.001$ \\
SPIRAL@400     & game play, multi-game  & $0.073$ & $0.755$ & $0.796$ & $\mathbf{0.787 \pm 0.005}$ & $0.822 \pm 0.001$ & $\mathbf{0.957 \pm 0.006}$ \\
\bottomrule
\end{tabular}
\end{table}

\begin{figure}[t]
\centering
\includegraphics[width=0.8\linewidth]{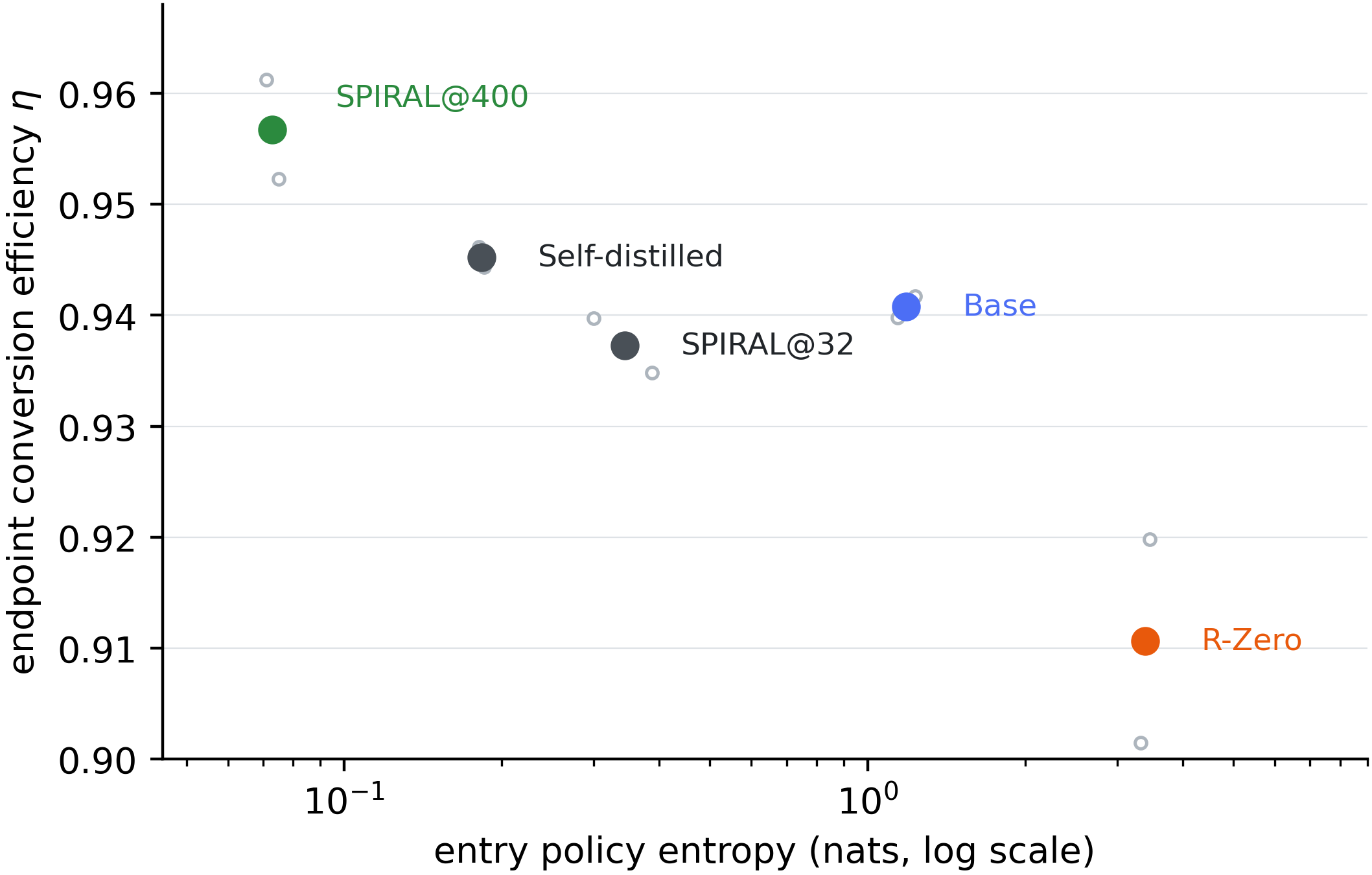}
\caption{Entry policy entropy versus endpoint conversion efficiency for the five data-free checkpoints. Filled markers denote seed means and hollow markers individual seeds $\{2,3\}$.}
\label{fig:scatter}
\end{figure}

Across the five data-free checkpoints, entry entropy strongly rank-orders endpoint conversion efficiency (Spearman $\rho=-0.900$, exact $p=0.083$). After rank-adjusting for entry best@16, the association strengthens to $\rho=-0.986$ ($p=0.050$), indicating that the relationship is not explained by differences in reachable capability. The small roster makes the raw two-sided permutation $p$ coarse (a five-point exact test cannot fall below $0.017$), so we read these as a strong, capability-controlled ordering rather than a powered significance test.

Conversion is also more stable than raw accuracy. Across seeds, checkpoint ranks by $\eta$ are identical ($\rho=1.000$, exact $p=0.017$), whereas final mean@16 ranks are not ($\rho=0.600$, $p=0.350$).

\textbf{The ordering replicates across domains.} Tables~\ref{tab:amc} and~\ref{tab:gpqa} repeat the protocol on GPQA and AMC under matched seeds ${2,3}$. GPQA provides a cross-domain test, while AMC probes a harder mathematics distribution. Entry entropy rank-orders endpoint $\eta$ at $\rho=-1.000$ on GPQA (exact $p=0.017$; $\rho=-1.000$ after adjusting for entry best@16) and $\rho=-0.800$ on AMC ($p=0.133$; partial $\rho=-0.728$). Thus the same direction holds both outside the math domain and under a harder in-domain shift, although the AMC result is directional rather than significant at $n=5$.

The checkpoint ordering is not identical across benchmarks. SPIRAL@400 finishes highest on MATH and GPQA, whereas the self-distilled checkpoint is marginally highest on AMC (final mean@16 $0.479$ vs.\ $0.476$; $\eta=0.684$ vs.\ $0.676$). We do not interpret this small difference as evidence for a universally superior preparation recipe; rather, it reinforces our focus on the entry state produced by preparation and its interaction with benchmark-specific reachability.

\begin{table}[t]
\centering
\caption{GPQA under the frozen 15-step TTRL protocol, matched seeds $\{2,3\}$, ordered by entry entropy. Entry columns are seed means; final columns and $\eta$ are means $\pm$ sample s.d.}
\label{tab:gpqa}
\scriptsize
\setlength{\tabcolsep}{3pt}
\begin{tabular}{llcccccc}
\toprule
\textbf{Checkpoint} & \textbf{Objective} & \textbf{Entry} & \textbf{Entry} & \textbf{Entry} &
\textbf{Final} & \textbf{Final} & \boldmath$\eta$ \\
& & \textbf{entropy} & \textbf{mean@16} & \textbf{best@16} &
\textbf{mean@16} & \textbf{best@16} & \\
\midrule
R-Zero         & uncert.\ curriculum    & $4.644$ & $0.305$ & $0.731$ & $0.317 \pm 0.006$ & $0.735 \pm 0.024$ & $0.432 \pm 0.021$ \\
Base           & --- (pretrained)       & $1.791$ & $0.239$ & $0.736$ & $0.357 \pm 0.013$ & $0.766 \pm 0.017$ & $0.466 \pm 0.007$ \\
SPIRAL@32      & game play, Kuhn @32    & $0.681$ & $0.349$ & $0.790$ & $0.383 \pm 0.004$ & $0.783 \pm 0.011$ & $0.489 \pm 0.012$ \\
Self-distilled & data-free, high-dose   & $0.454$ & $0.272$ & $0.758$ & $0.369 \pm 0.001$ & $0.727 \pm 0.005$ & $0.508 \pm 0.002$ \\
SPIRAL@400     & game play, multi-game  & $0.130$ & $0.383$ & $0.780$ & $\mathbf{0.406 \pm 0.002}$ & $0.755 \pm 0.010$ & $\mathbf{0.538 \pm 0.005}$ \\
\bottomrule
\end{tabular}
\end{table}

\begin{table}[t]
\centering
\caption{AMC under the frozen 15-step TTRL protocol, matched seeds $\{2,3\}$, ordered by entry entropy. Entry columns are seed means; final columns and $\eta$ are means $\pm$ sample s.d.}
\label{tab:amc}
\scriptsize
\setlength{\tabcolsep}{3pt}
\begin{tabular}{llcccccc}
\toprule
\textbf{Checkpoint} & \textbf{Objective} & \textbf{Entry} & \textbf{Entry} & \textbf{Entry} &
\textbf{Final} & \textbf{Final} & \boldmath$\eta$ \\
& & \textbf{entropy} & \textbf{mean@16} & \textbf{best@16} &
\textbf{mean@16} & \textbf{best@16} & \\
\midrule
R-Zero         & uncert.\ curriculum    & $3.452$ & $0.417$ & $0.710$ & $0.414 \pm 0.003$ & $0.665 \pm 0.024$ & $0.623 \pm 0.018$ \\
Base           & --- (pretrained)       & $1.041$ & $0.393$ & $0.739$ & $0.460 \pm 0.006$ & $0.747 \pm 0.001$ & $0.615 \pm 0.007$ \\
SPIRAL@32      & game play, Kuhn @32    & $0.278$ & $0.426$ & $0.684$ & $0.460 \pm 0.013$ & $0.725 \pm 0.004$ & $0.635 \pm 0.021$ \\
Self-distilled & data-free, high-dose   & $0.221$ & $0.403$ & $0.741$ & $\mathbf{0.479 \pm 0.006}$ & $0.700 \pm 0.002$ & $\mathbf{0.684 \pm 0.011}$ \\
SPIRAL@400     & game play, multi-game  & $0.076$ & $0.436$ & $0.744$ & $0.476 \pm 0.011$ & $0.704 \pm 0.009$ & $0.676 \pm 0.007$ \\
\bottomrule
\end{tabular}
\end{table}

Taken together, these results identify entry entropy as a measurable signature of TTRL readiness: before any adaptation occurs, it rank-orders how efficiently checkpoints convert reachable capability into reliable output, beyond differences in entry reachability and across multiple benchmarks.

\subsection{Data-free training effectively moves the entry state}
\label{sec:preparation}

The preceding result would be less useful if entry state were merely a fixed property of pretraining. Instead, different data-free objectives move the same base family to very different states.

R-Zero and SPIRAL@400 provide the clearest contrast. R-Zero's curriculum repeatedly selects problems on which the Solver disagrees with itself, whereas SPIRAL trains on discrete game outcomes verified by the environment. Consistent with these objectives, R-Zero enters at $3.391$ nats while SPIRAL@400 enters at $0.073$. Both are data-free, yet they produce opposite TTRL substrates. The relevant property is therefore not data-free training itself, but the entry state induced by the training objective.

Our in-domain self-distillation experiment provides a direct sharpening intervention. Starting from Base, label-free self-distillation moves entry entropy from $1.186$ to $0.183$ nats and increases endpoint conversion efficiency from $0.941$ to $0.945$. This shows that the entry state can be deliberately shifted in the direction predicted by the cross-checkpoint relationship.

At this high preparation strength, however, final reachability is lower than Base ($0.811$ vs.\ $0.820$), leaving final mean@16 marginally lower. This marks the constraint on sharpening. Entry entropy should therefore be treated as a calibration signal rather than an objective to minimize: preparation is useful insofar as it improves conversion without sacrificing reachable capability.

The two self-play checkpoints add complementary evidence that greater self-play concentrates the entry state further. SPIRAL@32, an early (step-32) endpoint of a Kuhn-Poker self-play run, enters at $0.344$ nats with $\eta=0.937$, whereas the completed multi-game SPIRAL@400 enters at $0.073$ nats with $\eta=0.957$. We see this as directional evidence that stronger self-play concentrates the policy and improves conversion.

\section{Limitations}
\label{sec:limitations}

Our main empirical result is observational across checkpoints with different training histories: entry entropy strongly predicts how efficiently short-horizon TTRL converts reachable capability, and data-free training can substantially move that entry state. A controlled within-method preparation-strength sweep is still required to establish the causal effect of calibrated entry-state sharpening on final TTRL performance.

Our experiments focus on the Qwen3-4B family and a fixed 15-step TTRL regime. MATH provides the strongest matched-seed evidence, while GPQA and AMC support the same directional pattern across additional domains. Whether the relationship generalizes across model families, adaptation horizons, and broader task distributions remains open.

Finally, we use mean token-level policy entropy as a practical proxy for entry-state concentration. This quantity is label-free and easy to measure, but it does not fully capture diversity at the answer or solution level; future work should test whether semantic or answer-level uncertainty provides a more direct measure of the reachable solution set.

\section{Conclusion}
\label{sec:conclusion}

We introduced \textbf{entry-state sharpening}: preparing a general-purpose checkpoint with data-free training before test-time reinforcement learning. The central premise is that TTRL largely converts reachable capability into reliable output, so part of its limited adaptation budget can be moved upstream by placing the model in a more concentrated entry state.

Across five data-free checkpoints derived from the same Qwen3-4B base, entry policy entropy strongly rank-orders endpoint conversion efficiency, including after controlling for entry reachability, and the same ordering replicates on GPQA and AMC. Importantly, these differences in conversion are reflected in end-task performance: appropriately prepared checkpoints can also finish above the untuned Base after TTRL, with SPIRAL achieving the highest final accuracy on MATH and GPQA and the self-distilled checkpoint marginally highest on AMC. The contrast between R-Zero and SPIRAL further shows that data-free training alone is not sufficient: different objectives can move the same base family toward sharply different entry states and correspondingly different TTRL outcomes. Finally, our in-domain self-distillation experiment shows that this state can be deliberately shifted, while also revealing the constraint on sharpening: improved conversion is useful only while reachable capability is preserved.

These results suggest a different way to prepare models for test-time learning. Rather than treating the checkpoint entering TTRL as fixed, we can treat its entry state as something to design. Entry entropy provides a simple label-free signal for that preparation; the remaining question is how strongly to sharpen a given checkpoint before reachability begins to deteriorate.

\bibliographystyle{plainnat}
\bibliography{references}

\end{document}